\documentclass[lettersize,journal]{IEEEtran}
\usepackage{amsmath,amsfonts}
\usepackage{algorithmic}
\usepackage{algorithm}
\usepackage{array}
\usepackage[caption=false,font=normalsize,labelfont=sf,textfont=sf]{subfig}
\usepackage{textcomp}
\usepackage{stfloats}
\usepackage{url}
\usepackage{verbatim}
\usepackage{graphicx}
\usepackage{cite}
\usepackage{xcolor}

\usepackage{hyperref}
\usepackage{multirow}

\begin{document}

\title{Emphasis Rendering for Conversational Text-to-Speech with Multi-modal Multi-scale Context Modeling}

\author{Rui Liu, \IEEEmembership{Member, IEEE}, Zhenqi Jia, Jie Yang, Yifan Hu and Haizhou Li  \IEEEmembership{Fellow, IEEE}

\thanks{
The research by Rui Liu was funded by the Young Scientists Fund (No. 62206136) and the General Program (No. 62476146) of the National Natural Science Foundation of China, Guangdong Provincial Key Laboratory of Human Digital Twin (No. 2022B121201 0004),
and the ``Inner Mongolia Science and Technology Achievement Transfer and Transformation Demonstration Zone, University Collaborative Innovation Base, and University Entrepreneurship Training Base'' Construction Project (Supercomputing Power Project) (No.21300-231510). 
The research by Haizhou Li was partly supported by Internal Project Fund from Shenzhen Research Institute of Big Data (Grant No.~T00120220002), and Shenzhen Science and Technology Research Fund (Fundamental Research Key Project Grant No.~JCYJ20220818103001002).
}

\thanks{Rui Liu, Zhenqi Jia, Jie Yang and Yifan Hu are with the Department of Computer Science, Inner Mongolia University, Hohhot 010021, China. 
(e-mail: liurui\_imu@163.com).}


\thanks{Haizhou Li is with School of Data Science, The Chinese University of Hong Kong, Shenzhen 518172, China. He is also with University of Bremen, Faculty 3 Computer Science / Mathematics, Enrique-Schmidt-Str. 5 Cartesium, 28359 Bremen, Germany (e-mail: haizhouli@cuhk.edu.cn). }

\thanks{Manuscript received Sep 29, 2024; revised August 16, 2021.}
}


\markboth{PREPRINT MANUSCRIPT OF IEEE/ACM TRANSACTIONS ON AUDIO, SPEECH, AND LANGUAGE PROCESSING}%
{Shell \MakeLowercase{\textit{et al.}}: Bare Demo of IEEEtran.cls for IEEE Journals}


\maketitle

\begin{abstract}
Conversational Text-to-Speech (CTTS) aims to accurately express an utterance with the appropriate style within a conversational setting, which attracts more attention nowadays. While recognizing the significance of the CTTS task, prior studies have not thoroughly investigated speech emphasis expression, which is essential for conveying the underlying intention and attitude in human-machine interaction scenarios, due to the scarcity of conversational emphasis datasets and the difficulty in context understanding. In this paper, we propose a novel Emphasis Rendering scheme for the CTTS model, termed ER-CTTS, that includes two main components: 1) we simultaneously take into account textual and acoustic contexts, with both global and local semantic modeling to understand the conversation context comprehensively; 2) we deeply integrate multi-modal and multi-scale context to learn the influence of context on the emphasis expression of the current utterance. Finally, the inferred emphasis feature is fed into the neural speech synthesizer to generate conversational speech. To address data scarcity, we create emphasis intensity annotations on the existing conversational dataset (DailyTalk). Both objective and subjective evaluations suggest that our model outperforms the baseline models in emphasis rendering within a conversational setting. The code and audio samples are available at https://github.com/CodeStoreTTS/ER-CTTS.
\end{abstract}

\begin{IEEEkeywords}
Emphasis, Speech Synthesis, Conversational TTS, Multi-scale, Multi-modal
\end{IEEEkeywords}

\section{Introduction}
\IEEEPARstart{C}{onversational} Text-to-Speech (Conversational TTS, CTTS) aims to synthesize speech with suitable prosody according to the conversation context \cite{guo2021conversational}. As Human-Computer Interaction (HCI) systems become more prevalent in our daily lives, CTTS garners significant attention \cite{tulshan2019survey,seaborn2021voice,mctear2022conversational}. We note that speech emphasis, an important prosodic feature in the dialogue process, plays an important role in distinguishing the focus of the utterance from the rest and conveys the underlying intention and attitude \cite{liu2021controllable}. In HCI scenarios, synthesizing emphasis helps computer systems to express semantics more accurately and enhances user experience, thus attracting increasing interest \cite{liu2021controllable}.

However, emphasis rendering is currently limited to the isolated utterance condition \cite{seshadri2021emphasis,raitio2022hierarchical} and lacks in-depth research in the context of CTTS \cite{guo2021conversational}. For example, Zhong et al. \cite{zhong2023ee} build an Emphatic Expressive TTS system by leveraging multi-level linguistic information from syntax and semantics to predict the emphasis position for the current utterance. Chi et al. \cite{chi2023multi} propose a multi-granularity, including phrase- and word-level, emphasis prediction model for TTS \cite{10379131,10487819}. There are also similar concepts, such as prosodic prominence \cite{cole2014listening, cole2017crowd, morrison2024crowdsourced}, stress \cite{chi2023multi, he2022automatic}, etc., that refer to certain words in a sentence being pronounced more prominently than others, but they do not take into account the relation between prominence and HCI. We note that these methods ignore the complex semantic dependencies between dialogue history and the emphasis expression of the current utterance. Therefore, how to understand the context information in a dialogue to make appropriate emphasis inferences for the current utterance, to achieve emphasis rendering for CTTS, is the focus of our paper.

Note that there are two challenges that currently pose difficulties in achieving this goal: 1) Multi-modal and multi-scale context. The dialogue history comprises both text and audio modalities \cite{xue2023m}. The text modality consists of a hierarchical semantic structure, ranging from sentence-level to word-level, and even phoneme-level \cite{lei2023msstyletts,shang2023hiertts}. Similarly, the audio modality follows a hierarchical acoustic structure, from sentence-level to frame-level \cite{shang2023hiertts}. Such multi-modal and multi-scale features play a crucial role in the expressiveness of the current utterance in conversation \cite{hu2022fctalker,xue2023m}. However, there is still a lack of clear conclusions and in-depth research on the specific impact of multi-modal and multi-scale features on the emphasis expression of the current sentence. 2) Emphatic CTTS dataset. Currently, some emphasis speech datasets only focus on isolated sentences \cite{zhong2023ee} and are not openly available. For emphasis speech datasets in conversational scenarios, there is a lack of a reasonable annotation scheme and readily available open-source data. Consequently, it becomes challenging to meet the research needs for emphasis rendering in CTTS.

In this work, we propose a new Emphasis Rendering scheme for CTTS, termed ER-CTTS, that consists of two novel modules. 1) To comprehensively understand the conversation context, we simultaneously model both textual and acoustic contexts with both coarse- and fine-grained modeling in the dialogue history. For all granularities of information, we adopt a bidirectional modeling approach to model the conversation history. In addition, for fine-grained information, we employ a memory enhancement method to capture historical emphasis cues. 2) To further effectively capture the implicit relationship between the rich context and the emphasis expression of the current utterance, we propose hybrid-grained and cross-modality fusion modules, which allow for the comprehensive fusion of context knowledge, facilitating the inference of emphasis intensity for the current utterance. Afterward, we incorporate the inferred emphasis feature into the FastSpeech2-based backbone to achieve emphasis rendering in CTTS.

To address the dataset gap, we propose an emphasis annotation scheme, specifically for conversation scenarios, on the popular DailyTalk \cite{lee2023dailytalk} dataset for CTTS. This annotation scheme allows for the acquisition of emphasis and emphasis intensity information. All the annotated data will be made open source. To summarize, we outline the main contributions of this work as follows: 
\begin{itemize}
\item We propose a novel emphatic conversational speech synthesis model, termed ER-CTTS. To our knowledge, this is the first in-depth conversational speech synthesis study that models emphasis rendering.
\item Our ER-CTTS aims to uncover implicit emphasis cues from multi-modal and multi-scale context information and effectively integrate this information to infer emphasis for the current utterance.
\item Objective and subjective experiments show that the proposed model outperforms all state-of-the-art baselines in terms of emphasis expressiveness.

\end{itemize}

\section{Related Works}
\subsection{Conversational Text-to-Speech}
The conventional CTTS works primarily focus on improving expressiveness through context understanding from various aspects, such as inter- and intra-speaker \cite{li2022enhancing}, and multi-modal context \cite{li2022enhancing,li2022inferring,xue2023m} dependencies modeling, emotion understanding \cite{liu2023emotion}, etc. The above research contributes to understanding the conversational context and determining appropriate speaking styles in synthesized speech. However, the explicit modeling of emphasis and its intensity in CTTS has not been extensively explored and requires further research.

In more recent studies, we also note some attempts to model the multi-modal and multi-scale context simultaneously. For example, the multi-scale relational graph convolutional network (MRGCN) proposed by \cite{li2022inferring} and the multi-scale multi-modal CTTS (M$^2$-CTTS) model proposed by \cite{xue2023m}, among others. However, our ER-CTTS has some clear differences from these works: 1) We develop a memory-enhanced fine-grained textual encoder to focus on local emphasis information in the dialogue history and avoid information loss; 2) Our context fusion encoder comprehensively integrates information from different granularities and modalities to make reasonable predictions about the emphasis information in the current utterance; 3) Our model learns more subtle emphasis intensity information within the dialogue history to assist in emphasis rendering for the final synthesized speech.

\begin{figure*}
  \centering
  \includegraphics[width=0.8\linewidth]{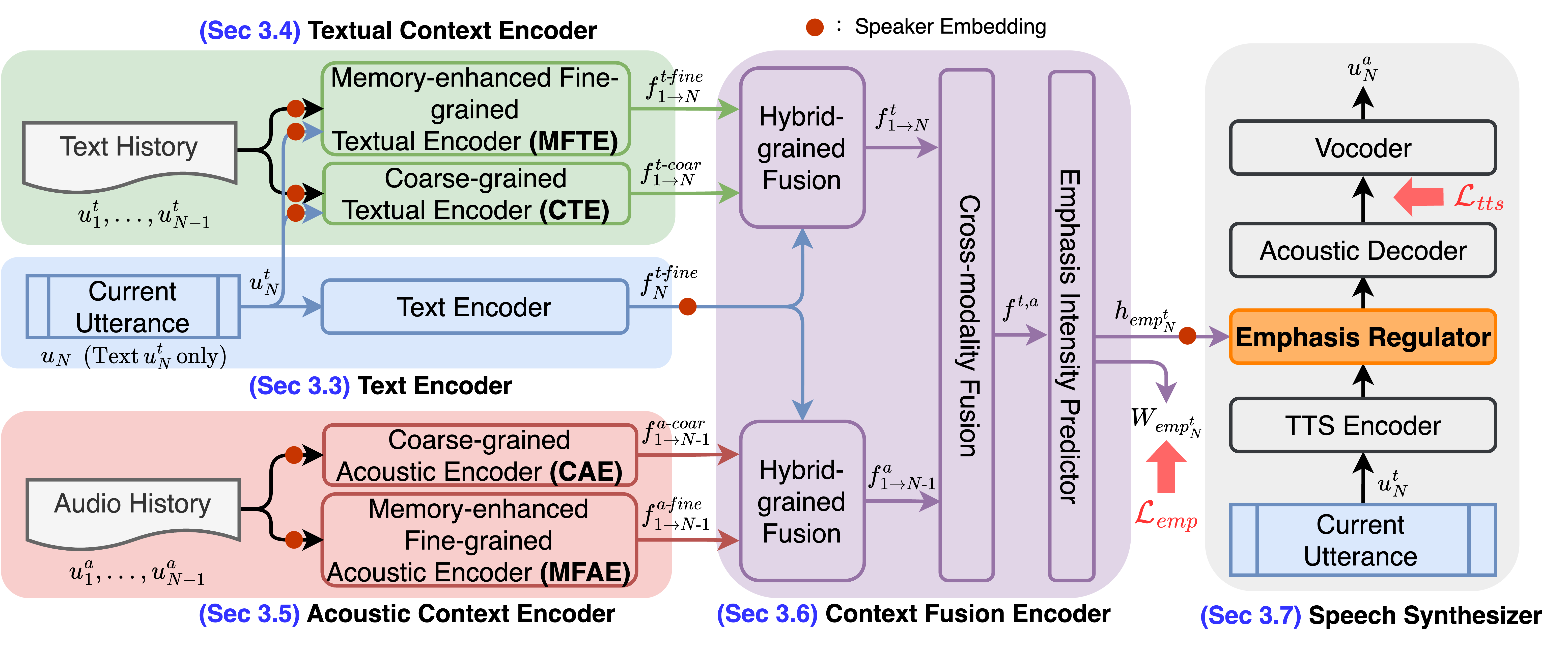}
  \caption{The overview of ER-CTTS, that consists of \textit{Text Encoder, Textual Context Encoder, Acoustic Context Encoder, Context Fusion Encoder} and the \textit{Speech Synthesizer}.}
  \label{fig:1}
\end{figure*}

\subsection{Conversational Dataset}
Many existing speech datasets for conversation focus on emotion recognition \cite{busso2008iemocap, poria2019meld, zhao2022m3ed}, but the audio quality often suffers from noise, making them less suitable for the data requirements of CTTS \cite{guo2021conversational}. While there are some datasets specifically recorded for CTTS, such as DailyTalk \cite{lee2023dailytalk}, MagicData-RAMC \cite{yang2022open}, etc., they lack explicit emphasis-related labels, which makes it challenging to model explicit emphasis rendering. In this work, we aim to construct an open-source emphasis rendering dataset for CTTS based on the DailyTalk dataset.

\section{ER-CTTS: Methodology}
\subsection{Task Definition}
A human-machine conversation can be defined as a sequence of utterances $U=\{[u_{1},s_{1}], [u_{2},s_{2}], ..., [u_{N-1},s_{N-1}], [u_{N},s_{N}]$\}, where \{$u_{1}, u_{2}, ..., u_{N-1}$\} is the sentence-level dialogue history till round $N-1$ while $u_{N}$ means the current utterance to be synthesized and $s_{i}$ ($i\in[1,N]$) is the speaker identity for each utterance.
Assuming that each utterance $u_{j}$ ($j\in[1,N-1]$) in the dialogue history is annotated with word-level emphasis information $W_{emp^{t}_j}$, achieving emphasis rendering in CTTS involves utilizing multi-modal information from the previous $N-1$ dialogue history to predict the emphasis information for the current utterance $u_{N}$. Note that $u_{j}$ includes $u^{t}_{j}$ and $u^{a}_{j}$ where $t$ and $a$ mean the text and acoustic modalities respectively. Additionally, the current utterance only consists of a text modality without the audio modality, as the audio signal is to be synthesized.

Subsequently, the conversational speech $u^{a}_{N}$ for $u_{N}$ is synthesized with the text input $u^{t}_{N}$ and the corresponding word-level emphasis information $W{{emp^{t}_{N}}}$. In other words, the emphasis expression of the synthesized speech $u^{a}_{N}$ should reflect the emphasis in the conversational context characterized by the multi-modal and multi-scale dialogue history. To this end, the emphasis rendering for CTTS methods needs to consider: 1) How to model the emphasis cues from the multi-modal and multi-scale conversation history; 2) How to integrate the knowledge of conversation history to infer the emphasis indicator for the current utterance; 3) How to render the appropriate emphasis expression during speech synthesis.

\subsection{Model Overview}
As shown in Fig. \ref{fig:1}, the proposed ER-CTTS consists of five components, which are 1) Text Encoder, 2) Textual Context Encoder, 3) Acoustic Context Encoder, 4) Context Fusion Encoder, and 5) Speech Synthesizer. The text encoder aims to generate the semantic encoding for the current utterance. The Textual and Acoustic Context Encoders seek to capture both textual and acoustic information from the conversation history, with utterance-level and phoneme-level context modeling. The Context Fusion Encoder is responsible for integrating the cross-grained and cross-modality interaction and inferring the emphasis expression for the current utterance. Finally, the Speech Synthesizer takes the current utterance and the emphasis expression as input to project the emphasis onto conversational speech \footnote{As shown by the red dots in Fig. \ref{fig:1}, the speaker embeddings corresponding to each sentence are fed into all the context encoders to distinguish speaker information. The speaker embedding of the current sentence is also fed into the emphasis regulator to synthesize the speech of the target speaker.}.

\subsection{Text Encoder}
To extract the high-level semantic representation of the word-level sequence $W_{u^{t}_{N}}$ of the current utterance $u^{t}_{N}$, we use XLNet \cite{yang2019xlnet} as the backbone to produce the fine-grained word-level feature $f^{t\text{-}\!fine}_{N}$.

\subsection{Textual Context Encoder}

To understand the dialogue context, with global and local views, by modeling the interaction between the history and the current utterance, the textual context encoder includes a \textbf{C}oarse-grained \textbf{T}extual \textbf{E}ncoder (\textbf{CTE}) and a \textbf{M}emory-enhanced \textbf{F}ine-grained \textbf{T}extual \textbf{E}ncoder (\textbf{MFTE}). 

\subsubsection{CTE}
The CTE is used to learn mutual dependencies \cite{ghosal2019dialoguegcn} between the dialogue history and the current utterance at the sentence level.
Therefore, we model the bidirectional dependency relationships to obtain the final coarse-grained textual context features.

Specifically, the dialogue textual history $u^{t}_{1},...,u^{t}_{N-1}$ and $u^{t}_{N}$ are fed to Sentence-BERT \cite{reimers2019sentence} to extract the sentence-level hidden vectors $h_{u^{t}_{1}},...,h_{u^{t}_{N-1}}$ and $h_{u^{t}_{N}}$, respectively. After that, unlike \cite{guo2021conversational}, which just uses GRU to encode the dialogue history, we employ bi-directional GRU (Bi-GRU) to encode the $h_{u^{t}_{1}},...,h_{u^{t}_{N-1}}$ into the state vector $h^{t-coar}_{N-1}$, representing the historical dialogue. Then, $h_{u^{t}_{N}}$ is concatenated with $h^{t-coar}_{N-1}$ and fed into a linear layer to obtain the interactive prosodic features as the final coarse-grained context feature $f^{t \text{-} coar}_{1 \rightarrow N}$.

\subsubsection{MFTE}
Inspired by real-world conversations, individuals tend to recall and review specific emphasized words from previous dialogues during the temporal discourse and respond accordingly \cite{liu2023emotionally} \footnote{We assume that the emphasis word annotation in the conversation history is known in this work. During the inference stage, the emphasis words in the conversation history are predicted using the same method to better predict the emphasis words in the current sentence.}. Therefore, MFTE is employed to remember and organize the emphasized information in the historical conversation, which also helps mitigate information loss during sequence learning and ensures that the important emphasis details are not forgotten.

\begin{figure*}
    \centering
    \includegraphics[width=1\linewidth]{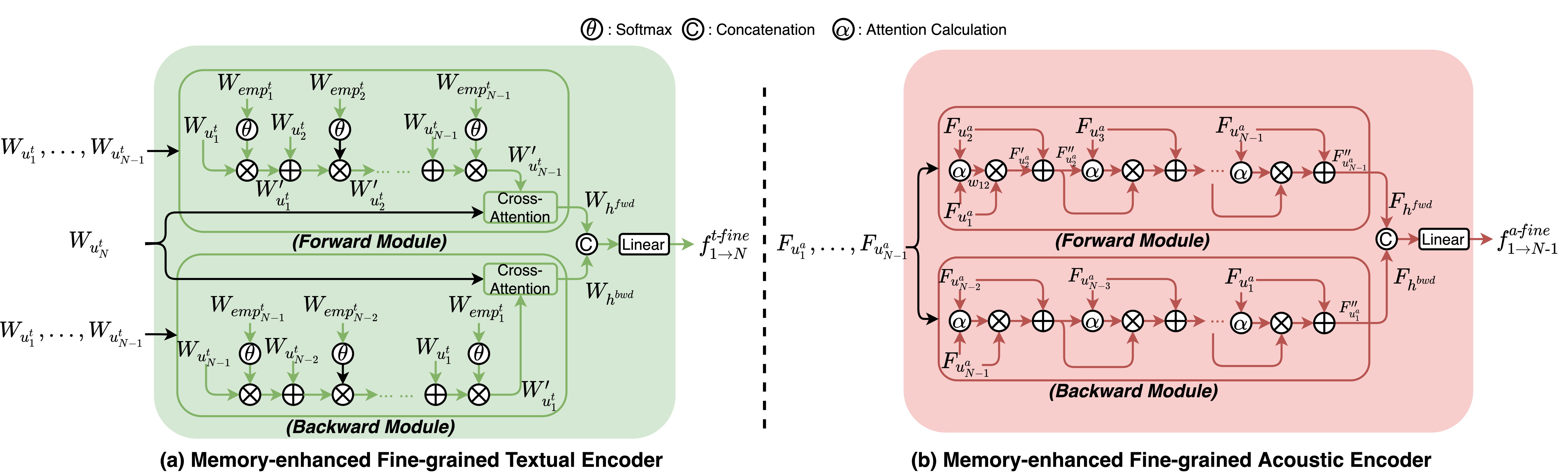}
    \caption{The detailed diagrams of \textit{Memory-enhanced Fine-grained Textual Encoder} and \textit{Memory-enhanced Fine-grained Acoustic Encoder}.}
     
    \label{fig:2}
\end{figure*}

The core idea of MFTE is to merge the word-level emphasis intensity value with the words of the current sentence. These combined features are then added to the words of the next sentence, and this process is repeated iteratively.
Firstly, we follow \cite{hu2022fctalker} and feed the sentence sequence $u^{t}_{1},...,u^{t}_{N}$ into the conversational language model, which is TOD-BERT \cite{wu-etal-2020-tod}, to extract word-level hidden vectors $W_{u^{t}_{1}},..., W_{u^{t}_{N}}$. The word-level dialogue history is represented as $W_{u^{t}_{1}},..., W_{u^{t}_{N-1}}$, and the current sentence is represented as $W_{u^{t}_{N}}$.
As shown in Fig. \ref{fig:2}(a), the MFTE consists of forward and backward modules. We note that the $W_{u^{t}_{1}},..., W_{u^{t}_{N-1}}$ and the $W_{u^{t}_{N-1}}$ are fed into these two modules to obtain forward and backward textual representations $W_{h^{fwd}}$ and $W_{h^{bwd}}$.

Specifically, taking the forward module as an example, the forward encoding process is divided into $N-1$ steps. For each step $j$, the emphasis information $W_{emp^{t}_{j}}$ is first combined with $W_{u^{t}_{j}}$ to output the enhanced representation $W'_{u^{t}_{j}}$ for $W_{u^{t}_{j}}$:

 \begin{equation}
    W'_{u^{t}_{j}} = Mul(W_{u^{t}_{j}}, Softmax(W'_{emp^{t}_{j}}))
\end{equation}
where ``Mul'' indicates the multiplication operation and the ``Softmax'' is used to smooth out word-level emphasis intensity to a range of 0 to 1.

The above process involves iterative computations until obtaining the output $W'_{u^{t}_{N-1}}$ of the $N-1$ step. Then we adopt cross-attention, which treats the $W'_{u^{t}_{N-1}}$ as the Key, Value and $W_{u^{t}_{N}}$ as the Query, to calculate the final forward context encoding $W_{h^{fwd}}$. Note that the reverse process follows the same computation method, starting from the $N-1$ step and continuing until the $1st$ step, resulting in the reverse encodings $W_{h^{bwd}}$.
At last, $W_{h^{fwd}}$ and $W_{h^{bwd}}$ are concatenated and fed into a linear layer to obtain the memory-enhanced context features as the final fine-grained textual context features $f^{t-fine}_{1 \rightarrow N}$.

\subsection{Acoustic Context Encoder}

The acoustic context encoder aims to understand the dialogue context with global and local views for acoustic, which is similar to the textual context encoder. However, the global and local views for acoustic refer to the sentence and speech frame levels, respectively. It consists of two modules that include \textbf{C}oarse-grained \textbf{A}coustic \textbf{E}ncoder (\textbf{CAE}) and \textbf{M}emory-enhanced \textbf{F}ine-grained \textbf{A}coustic \textbf{E}ncoder (\textbf{MFAE}).

\subsubsection{CAE}
Similar to the coarse-grained textual encoder, the coarse-grained Acoustic Encoder also obtains the final coarse-grained acoustic context features by considering bidirectional dependency modeling. Specifically, we follow \cite{xue2023m} and adopt a pre-trained Wav2vec 2.0 \cite{baevski2020wav2vec} fine-tuned on IEMOCAP \cite{busso2008iemocap} to extract the sentence-level prosody feature. Then, we also encode the sentence-level prosody feature using Bi-GRU to obtain the $f^{a-coar}_{1 \rightarrow N-1}$ as the coarse-grained acoustic context feature.

\subsubsection{MFAE}
Unlike the MFTE, the MFAE directly accumulates frame-level prosodic features, learning acoustic rhythm knowledge from the dialogue history. Note that we also adopt a bidirectional memory-enhanced structure, which includes the forward and backward modules, similar to the MFTE, as shown in Fig. \ref{fig:2}(b).

The core idea is to capture the prosodic interaction information between the previous and current utterances and then append it to the current utterance as its representation. Firstly, we follow \cite{xue2023m} and employ Wav2vec 2.0 \cite{baevski2020wav2vec} to extract frame-level acoustic features $F_{u^{a}_{1}},...,F_{u^{a}_{N-1}}$.

Take the forward module as an example; it reads the $F_{u^{a}_{1}},...,F_{u^{a}_{N-1}}$ and is also divided into $N-2$ steps. For step $k$, the frame-level representations of the previous sentence $F_{u^{a}_{k-1}}$ and the current sentence $F_{u^{a}_{k}}$ are treated as the Query, Key, and Value, respectively, to compute the attention weights $w_{k-1,k}$ at the frame level. These attention weights are then combined with the representation of the previous sentence $F_{u^{a}_{k-1}}$ to obtain $F'_{u^{a}_{k}}$. Subsequently, $F'_{u^{a}_{k}}$ is fused with the representation $F_{u^{a}_{k}}$ of the current sentence to obtain the final representation of the current sentence $F''_{u^{a}_{k}}$. Then, $F''_{u^{a}_{k}}$ will be computed with the next sentence $F_{u^{a}_{k+1}}$ in the same way, and this process continues until reaching the final $N-1$ step, outputting the final $F''_{u^{a}_{N-1}}$ as the final output $F_{h^{fwd}}$ of the forward module.

Similarly, we start from the $N-1$ step and perform the same computation with the history sentence, iteratively continuing until reaching the 1st utterance. This process generates the final output $F_{h^{bwd}}$ of the backward module. Subsequently, $F_{h^{fwd}}$ and $F_{h^{bwd}}$ are concatenated and fed into a linear layer to obtain the memory-enhanced frame-level prosodic features, serving as the final fine-grained acoustic context features $f^{a-fine}_{1 \rightarrow N-1}$.

\subsection{Context Fusion Encoder}
Context Fusion Encoder aims to comprehensively understand the emphasis cues from different granularities and modalities in the dialogue history and infer the emphasis intensity for the current utterance.

\subsubsection{Hybrid-grained Fusion}
 
Hybrid-grained Fusion first fuse coarse-grained and fine-grained contextual features within each modality.
1) For the text modality, we first repeat the sentence-level context features $f^{t-coar}_{1\rightarrow N}$ to each word in the current utterance, and add them to the word-level feature $f^{t-fine}_{N}$ to enable each word to focus on global semantic features, termed $f^{t-gbl}$. To integrate finer textual semantic emphasis cues, we treat the $f^{t-gbl}$ as the Query and  $f^{t-fine}_{1->N}$ as the Key and Value, and obtain the fusion feature $f^{t}_{1 \rightarrow N}$  with the cross-attention mechanism. 2) For the audio modality, we also first extend the coarse-grained context acoustic features $f^{a-coar}_{1\rightarrow N-1}$ and assign it to the word-level feature $f^{t-fine}_{N}$ to obtain the enhanced feature $f^{a-gbl}$, allowing each word to focus on global prosodic features.
We then treat $f^{a-gbl}$ as the Query and $f^{a-fine}_{1->N-1}$ as the Key and Value, and input into the cross-attention mechanism to obtain the fused frame-level prosodic cues, resulting in the feature $f^{a}_{1 \rightarrow N-1}$.

\subsubsection{Cross-modality Fusion}
Cross-modality Fusion aims to integrate multi-modal knowledge to understand the emphasis cues in a dialogue history.  Specifically, we treat the fused text contextual features $f^{t}_{1 \rightarrow N}$ as the Query, and the fused audio contextual features $f^{a}_{1 \rightarrow N-1}$ as the Key and Value to calculate the final integrated contextual features $f^{t,a}$ which combine both multi-modal and multi-scale aspects of the dialogue history. This design is intended to better capture the correlation between the text and audio modalities, thereby improving the accurate prediction of emphasis intensity.

\subsubsection{Emphasis Intensity Predictor}
The emphasis intensity predictor reads the fused context features $f^{t,a}$ to predict the emphasis intensity value $W_{emp^{t}_{N}}$ for each word of the current utterance $u^{t}_{N}$.
For the emphasis prediction loss $\mathcal{L}_{emp}$, we adopt the binary cross entropy loss to close the gap between the ground truth word-level emphasis information $W'_{emp^{t}_{N}}$ and the predicted label $W_{emp^{t}_{N}}$:
\begin{equation}
\begin{split}
    \mathcal{L}_{emp} = - \Big( & W'_{emp^{t}_{N}} \cdot \log(W_{emp^{t}_{N}}) \\
    & + (1 - W'_{emp^{t}_{N}}) \cdot \log(1 - W_{emp^{t}_{N}}) \Big)
\end{split}
\end{equation}

\subsection{Speech Synthesizer}
As shown in the right panel in Fig. \ref{fig:1}, the Speech Synthesizer takes FastSpeech2 as the backbone. It consists of TTS encoder, Acoustic Decoder and the Vocoder. TTS encoder aims to extract the phoneme-level linguistic encoding $P_{u^{t}_{N}}$ for current utterance $u^{t}_{N}$. The Acoustic Decoder involves a length regulator and variance adaptor to predict duration, energy, and pitch, and then adopts the Mel Decoder to predict Mel-spectrum features. Finally, a pre-trained HiFi-GAN Vocoder is used to generate the conversational speech $u^{a}_{N}$ with the desired emphasis expression.

Note that we include a new Emphasis Regulator on the top of TTS encoder to inject the emphasis knowledge for the linguistic encoding.
Unlike previous approaches \cite{chi2023multi} that directly inject one-hot emphasis label, we follow \cite{seshadri2021emphasis} and utilize hidden representation, that captures word-level emphasis information to achieve a more robust emphasis rendering effect.
. Specifically, we first extract the word-level hidden emphasis feature $h_{emp^{t}_{N}}$ from the last MLP layer of the emphasis intensity predictor.
Emphasis Regulator takes the $h_{emp^{t}_{N}}$ along with the $P_{u^{t}_{N}}$ as inputs.
We follow the word-phoneme alignment and integrate them into phoneme-level emphasis-enhanced feature representations, that are fed into subsequent acoustic decoder and vocoder to generate the final emphatic speech $u^{a}_{N}$.

The speech generation loss $\mathcal{L}_{tts}$ just follows the same style of FastSpeech2 \cite{ren2020fastspeech} to ensure that the synthesized speech is natural and close to ground truth speech.

\begin{figure}[h!]
  \centering
  \includegraphics[width=1\linewidth]{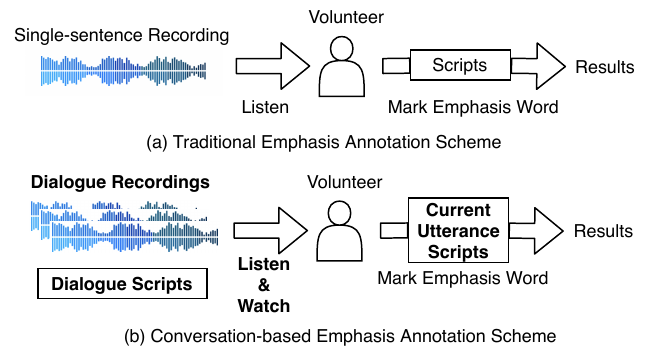}
  \caption{Conversation-based Emphasis Annotation Scheme.}
  \label{fig:3}
\end{figure}

\section{Dataset Annotation}

To meet our ER-CTTS training, we construct a new emphasis-aware conversational speech corpus based on the DailyTalk dataset, termed Emp-DailyTalk. We invite six volunteers to annotate emphasized words. Traditional emphasis annotation schemes, as shown in Fig. \ref{fig:3}(a),  often involve volunteers listening to a single sentence's audio and annotating the word-level emphasis in the corresponding text scripts \cite{mass2018word}. In contrast, our conversation-based emphasis annotation scheme, as shown in Fig. \ref{fig:3}(b), requires volunteers to first listen to the audio of the entire dialogue history and read the text of the dialogue history. Then they annotate the emphasized words in the current sentence text scripts. This approach considers the influence of both the semantic and acoustic features of the dialogue history on the current emphasis annotation, resulting in more accurate emphasis annotation. The detailed emphasis annotation scheme is as follows:

\subsubsection{Conversation-based Emphasis Annotation Scheme}

We adopt the IO scheme \cite{sang2003introduction}, using a set of dialogues as a unit and labeling each word in every utterance within the dialogues with “I" or “O". “I” represents the emphasis words, while “O” means the non-emphasis words. 
We invite six professional English graduate students with expertise in English language skills, including listening, speaking, reading, and writing, to annotate the data. Each sentence is annotated by all six volunteers, resulting in six emphasis annotations for each sentence. To further obtain the subtle emphasis intensity for each utterance, we follow \cite{shirani2019learning} and calculate the emphasis intensity label for each word in the entire sentence by taking the count of the “I” label among the six annotations and dividing it by 6. This provides us with the emphasis intensity label for each word. Table \ref{t-tab1} shows an example.

\begin{table}[h!]
\centering

\caption{An example of the emphasis annotation and the emphasis intensity calculation. ``A'' means the annotator.}

\begin{tabular}{c|c|c|c|c|c|c|c}
\hline
\textbf{Words} & \textbf{A1} & \textbf{A2} & \textbf{A3} &\textbf{ A4} & \textbf{A5} & \textbf{A6} & \textbf{Emphasis Intensity} \\
\hline
What & O & O & O & O & O & O & 0  \\
are & O & O & O & O & O & O & 0 \\
you & O & O & O & O & O & O & 0 \\
working & I & O & I & I & I & I & 0.83 \\
on & O & O & O & O & 1 & O & 0.17  \\
\hline
\end{tabular}

\label{t-tab1}
\end{table}

\subsubsection{Annotation Statistics}
Using the aforementioned annotation method, we perform emphasis annotation on 23,773 sentences from the Dailytalk dataset, resulting in emphasis annotation information for 23,773 sentences. Assuming a threshold of 0.5 for emphasis intensity, words with a value greater than 0.5 are considered emphasized words, while those with a value below 0.5 are considered non-emphasized words. Ultimately, the percentage of emphasized words in the entire dataset is approximately 23.76\%. We partition the data into training, validation, and test sets at a ratio of 7:2:1.

\subsubsection{Availability Verification}

We validate the consistency of emphasis annotations by calculating the kappa coefficient \cite{fleiss2013statistical}, resulting in a value of 0.35. In our study, as shown in Table \ref{t-tab2}, \cite{mass2018word} reports a kappa coefficient of 0.35 for emphasis annotations, while \cite{morrison2024crowdsourced} reports a kappa coefficient of 0.226. Despite some differences between emphasis annotations and emotion annotations (emphasis annotations focus on the word level, while emotion annotations focus on the sentence level), we still compare our results with emotion speech datasets. For example, the Emotionlines dataset achieves a kappa score of 0.34, slightly lower than our result. This evidence suggests that our data annotations are highly reliable, reaching a level comparable to or even higher than publicly available datasets. 

\begin{table}[h]
\centering
\caption{Comparison of Kappa coefficients between other datasets and Our Emp-DailyTalk.}

\resizebox{0.4\textwidth}{!}{
\begin{tabular}{c|c}
\hline
\textbf{Dataset} & \textbf{Kappa Coefficient} \\
\hline
Emotionlines \cite{chen2019emotionlines} & 0.340 \\
Emphasis Dataset 1 \cite{mass2018word} & 0.350 \\
Emphasis Dataset 2 \cite{morrison2024crowdsourced} & 0.226 \\
\hline
\textbf{Emp-DailyTalk} & 0.350 \\
\hline
\end{tabular}
}

\label{t-tab2}
\end{table}

\subsubsection{Crowdsourced Annotation Instructions}
\begin{figure}
    \centering
    \includegraphics[width=0.75\linewidth]{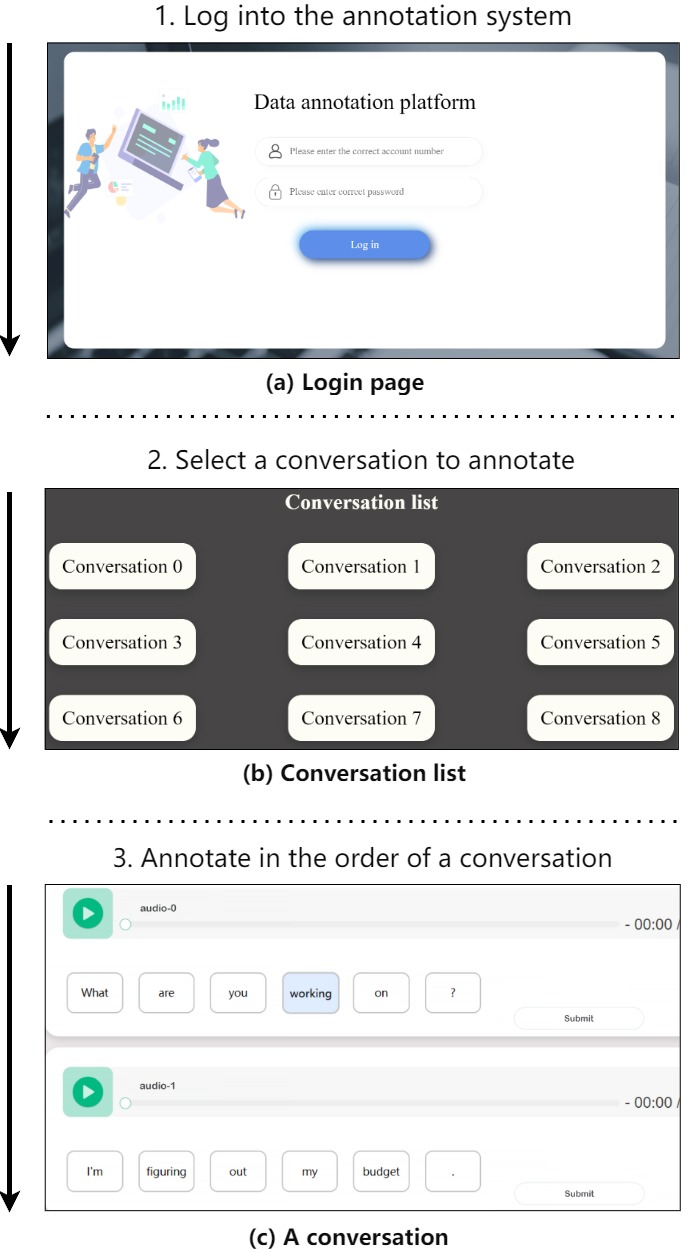}
    \caption{The annotation process of the Emp-DailyTalk.}
    \label{fig:4}
\end{figure}

Speech emphasis, an essential prosodic feature in dialogue, plays a pivotal role in distinguishing the focal points of utterances, conveying underlying intentions, and expressing attitudes. Prior to annotating the emphasized words in a sentence, volunteers are required to consider both the textual and audio components of the dialogue history.
As depicted in Fig. \ref{fig:4} (a), volunteers begin by logging into the annotation platform. They then proceed to select a conversation from the conversation list, as shown in Fig. \ref{fig:4}(b). The text and audio of the selected conversation are displayed in the order of the conversation, as illustrated in Fig. \ref{fig:4}(c). Volunteers are required to annotate the emphasis in the sequence of the conversation.
Volunteers receive their annotation tasks online. To ensure fair compensation, we use a unit of ten sentences for each task. Annotating ten sentences typically takes around 10 minutes. We offer a compensation of \$3 for each completed unit, which amounts to an estimated \$18 per hour.





\section{Experiments}

\subsection{Comparative Systems}
To validate the ER-CTTS from a more comprehensive perspective, we categorize all baselines into the following four classes:

\textbf{(1) Single-sentence TTS without emphasis:} \textbf{FastSpeech2} \cite{ren2020fastspeech} is a TTS model without emphasis and contextual modeling, representing state-of-the-art non-dialogue TTS systems.

\textbf{(2) Single-sentence TTS with emphasis:} \textbf{FastSpeech2 w/ Emphasis} \cite{seshadri2021emphasis} focuses on synthesizing emphasis speech for individual sentences. It leverages FastSpeech2 as the backbone and is studied to change the degree of emphasis by adjusting specific acoustic features, such as pitch and energy.

\textbf{(3) CTTS without emphasis:} \textbf{DailyTalk} \cite{lee2023dailytalk} is an advanced CTTS baseline. It proposes coarse-grained text context modeling in dialogue history to enhance speech expressiveness. \textbf{FCTalker} \cite{hu2022fctalker} further employs coarse-grained and fine-grained context modeling. \textbf{M$^2$-CTTS} \cite{xue2023m} designs a textual context module and an acoustic context module with both coarse-grained and fine-grained modeling. \textbf{GCN} \cite{li2022inferring} adopts the homogeneous graph to model the multi-modal context in conversation.

\textbf{(4) Expressive CTTS:} \textbf{ECSS} \cite{liu2023emotion} is a powerful expressive conversational TTS. It utilizes heterogeneous graph-based context modeling to achieve expressive rendering for CTTS.

\subsection{Evaluation Metrics}
\textbf{Emphasis Prediction:} To evaluate the performance of emphasis prediction objectively, we follow \cite{shirani2019learning} and adopt $\text{Match}{m}$ \cite{shirani2019learning} and $\text{F1}{m}$ \cite{shirani2019learning} as the accuracy metrics. We set $m={1,2}$. Both metrics are higher the better.
\textbf{Emphasis Rendering:}
To validate the emphasis rendering for ER-CTTS, we follow \cite{liu2023emotion} and conduct dialogue-level Mean Opinion Score (DMOS) \cite{streijl2016mean} as the subjective evaluation metric. Note that in this paper, DMOS consists of natural DMOS (N-DMOS) and emphasis DMOS (E-DMOS), where N-DMOS focuses on speech naturalness while E-DMOS assesses whether the emphasis expression in terms of position and intensity of the current utterance matches the context. The detailed scoring criteria of E-DMOS are shown in Table \ref{t-tab3} of Appendix \ref{app
}. For the objective evaluation metrics, we calculate the Mean Absolute Error (MAE) between the predicted and real acoustic features. Specifically, we assess the acoustic feature in terms of pitch, energy, and duration with MAE-P, MAE-E, and MAE-D.

\textbf{$\text{Match}_{m}$}  For each instance u in the test set $D_{test}$, we select a set $G_m^{(u)}$ of $m \in \lbrace 1, 2 \rbrace$ words with the top $m$ probabilities according to the ground truth. Analogously. We select a prediction set $\hat{G}_m^{(u)}$ for each $m$, based on the predicted probabilities. We define the metric $\text{Match}_{m}$ as follows:

\begin{equation}
\begin{split}
    \operatorname{Match}_m=\frac{\sum_{u \in D_{\text {test }}}\left|G_m^{(u)} \cap \hat{G}_m^{(u)}\right| / \min (m,|u|)}{\left|D_{\text {test }}\right|}
\end{split}
\end{equation}

\textbf{$\text{F1}_{m}$}  Similarly to  $\text{Match}_{m}$, for each instance $u$, we select the top $m$ = $\lbrace 1, 2 \rbrace$ words with the highest probabilities from both ground truth and prediction. Then F1-score per each $m$ can be computed accordingly.

\textbf{$\text{E-DMOS}$} scoring criteria:
As shown in Table \ref{t-tab3}, the volunteers refer to the dialogue history of the current sentence and evaluate the alignment of emphasis position and intensity with the dialogue contextual, on a scale of 1 to 5.

\begin{table}[h]
\centering

\caption{E-DMOS subjective evaluation scoring criteria.}
\label{t-tab3}

\begin{tabular}{c|c}
\hline
\textbf{Score} & \textbf{Criteria} \\
\hline
5  & Emphasis position and intensity are perfectly appropriate \\
4  & Emphasis position and intensity are reasonable \\
3  & Emphasis position is reasonable but intensity is unreasonable \\
2  & Emphasis position and intensity are unreasonable \\
1  & Emphasis position and intensity are completely inappropriate \\
\hline
\end{tabular}

\end{table}

\subsection{Ablation Systems}
We utilize the following ablation methods to assess the contributions offered by different components of ER-CTTS. Abl.Exp.1-7 are used to verify the contributions of different components to Emphasis Rendering, while Abl.Exp.8-20 are used to verify the contributions of different components to Emphasis Prediction.

\textbf{Abl.Exp.1\&2}: \textit{w/o Coarse-grained Encoders (CTE \& CAE)} and \textit{w/o Fine-grained Encoders (MFTE \& MFAE}) mean we eliminate two coarse-grained and two fine-grained encoders respectively;

\textbf{Abl.Exp.3\&4}: \textit{w/o Hybrid-grained Fusion} and \textit{w/o Cross-modality Fusion} indicate that we replace those two fusion modules with simple feature addition;

\textbf{Abl.Exp.5}: \textit{w/o Bidirectional Context Modeling} means we remove the bidirectional modeling mechanism from the context encoders and replace it with forward modeling only; 

\textbf{Abl.Exp.6}: \textit{w/o Memory Enhancement} solely employs feature concatenation for accumulating historical information.; 

\textbf{Abl.Exp.7}: \textit{w/o Emphasis Intensity} represent emphasis features as 0/1 labels instead of intensity features. 

\textbf{Abl.Exp.8-16}: exhaustively explores all possible combinations of the four encoders, including CTE, MFTE, CAE and MFAE.

\textbf{Abl.Exp.17-20}: are similar to Abl.Exp.3-6, which also involve removing the respective modules. However, the difference is that Abl.Exp.17-20 focus only on evaluating emphasis intensity prediction and do not consider emphasis rendering.

\subsection{Implementation Details}
For Textual encoders, the dimensions of $h_{u^{t}_{1}},...,h_{u^{t}_{N-1}}$, $W_{u^{t}_{1}},...,W_{u^{t}_{N-1}}$ and $W_{u^{t}_{N}}$ are set to 512, 768 and 1024 respectively. 
For Acoustic encoders, the dimensions of $h_{u^{a}_{1}},...,h_{u^{a}_{N-1}}$ and $F_{u^{a}_{1}},...,F_{u^{a}_{N-1}}$ are set to 768.
All speaker embeddings $s_{1}, s_{2}, ..., s_{N}$ are mapped to a vector with 768-dimension.
In all coarse-grained encoders, the output dimension of the Bi-GRU is 128, and the GRU layer is set to 2. The final output dimension is 256. For all two memory-enhanced fine-grained encoders, the number of attention heads in the attention module is 3. The dimensions of Query, Key, and Value are all 768. 
The input dimension of the Linear layer is 768, and the output dimension is 256.
In the context fusion encoder, the number of attention heads is set to 2, and the dimensions of Query, Key, and Value are all 256.
The Emphasis Intensity Predictor consists of two MLP layers with \{256,64,1\} dimension setup. For speech synthesizer, we employ the Adam optimizer with $\beta$1 = 0.9 and $\beta$2 = 0.98. Grapheme-to-Phoneme (G2P) toolkit \footnote{\href{https://github.com/Kyubyong/g2p}{https://github.com/Kyubyong/g2p}} is used for converting all text inputs into their respective phoneme sequences.
We utilize the Montreal Forced Alignment (MFA) \cite{mcauliffe17_interspeech} tool to extract phoneme duration alignment.
All speech samples are re-sampled to 22.05 kHz. Mel-spectrum features are extracted with a window length of 25ms and a shift of 10ms. The model is trained on an A100 GPU with a batch size of 16 for 900k steps. 


\begin{table*}[t!]
\centering

\caption{\label{t-tab4}\textcolor{black}{\textcolor{black}{Subjective (with 95\% confidence interval
) }and objective results for comparative and ablation experiments in terms of emphasis rendering. (Bold text indicates the best performance in the comparative experiment, while * denotes the suboptimal performance in the ablation results.)}} 

\resizebox{1\linewidth}{!}{
\begin{tabular}{lcccccccccccc}
\hline
\textbf{Systems}   &     & \textbf{N-DMOS} ($\uparrow$)&    & \textbf{E-DMOS} ($\uparrow$)            &                    & \multicolumn{1}{l}{\textbf{MAE-P ($\downarrow$)}} &    & \multicolumn{1}{l}{\textbf{MAE-E ($\downarrow$)}}    &  & \multicolumn{1}{l}{\textbf{MAE-D ($\downarrow$)}}     & \\ \hline

 
 FastSpeech2 \cite{ren2020fastspeech}  & & 3.602 $\pm$ 0.017 & & 3.559 $\pm$ 0.016 & & 0.645 & & 0.558 & & 0.240 & \\

 FastSpeech2 w/ Emphasis \cite{seshadri2021emphasis}  & & 3.636 $\pm$ 0.019&  & 3.664 $\pm$ 0.017 &  & 0.622 & & 0.581 & & 0.240&  \\

 DailyTalk \cite{lee2023dailytalk} &  & 3.637 $\pm$ 0.021 & & 3.647 $\pm$  0.021 & & 0.395 & & 0.282 & & 0.183 & \\
 
 FCTalker \cite{hu2022fctalker} &  & 3.686 $\pm$ 0.016&  & 3.683 $\pm$ 0.016&  & 0.397 & & 0.282 & & 0.181 & \\

 \textcolor{black}{M$^2$-CTTS} \cite{xue2023m} &  & 3.716 $\pm$ 0.015&  & 3.705 $\pm$ 0.016&  & 0.433 & & 0.200 & & 0.157 & \\

 GCN \cite{li2022inferring}  & & 3.725 $\pm$ 0.018 & & 3.704 $\pm$ 0.019& & 0.456 & & 0.204&  & 0.150 &  \\

 ECSS \cite{liu2023emotion}  & & 3.749 $\pm$ 0.019 & & 3.713 $\pm$ 0.019 & & 0.455&  & 0.215 & & 0.152&  \\
 
\hline
\textbf{ER-CTTS (Proposed)} & & \textbf{3.852 $\pm$ 0.018} & & \textbf{3.893 $\pm$ 0.023}  & & \textbf{0.342} & & \textbf{0.199*} & & \textbf{0.148} & \\
\hline

\textit{Abl.Exp.1}: w/o Coarse-grained Encoders (CTE \& CAE) &  & 3.788 $\pm$ 0.016 & & 3.798 $\pm$ 0.023&  & 0.398 & & 0.176 & & 0.169 & \\

\textit{Abl.Exp.2}: w/o Fine-grained Encoders (MFTE \& MFAE) &  & 3.693 $\pm$ 0.022 & & 3.720 $\pm$ 0.020&  & 0.458 & & 0.497 & & 0.210&  \\

\textit{Abl.Exp.3}:  w/o Hybrid-grained Fusion  & & 3.752 $\pm$ 0.017 & & 3.761 $\pm$ 0.017&  & 0.423 & & 0.235 &&  0.178 & \\
 
\textit{Abl.Exp.4}: w/o Cross-modality Fusion  & & 3.766 $\pm$ 0.016 & & 3.809 $\pm$ 0.024&  & 0.401 & & 0.188 & & 0.157&  \\
\hline
\textit{Abl.Exp.5}: w/o Bidirectional Context Modeling &  & 3.766 $\pm$ 0.018 & & 3.775 $\pm$ 0.012 & & 0.402 & & 0.214 &&  0.165&  \\

\textit{Abl.Exp.6}: w/o Memory Enhancement  & & 3.703 $\pm$ 0.019 & & 3.705 $\pm$ 0.018  &&  0.396 & & 0.182 & & 0.176 & \\

\textit{Abl.Exp.7}: w/o Emphasis Intensity  &  & 3.756 $\pm$ 0.015 & & 3.735 $\pm$ 0.028 &  & 0.403 & & 0.197 & & 0.182 & \\
\hline
\end{tabular}
}

\end{table*}

\begin{table*}[t!]
\centering

\caption{\label{t-tab5}The ablation results in terms of emphasis intensity prediction. ``-'' indicates the removal of a particular module, while ``$\checkmark$'' indicates its retention.}

\resizebox{1\linewidth}{!}{
\begin{tabular}{l|cccc|cccccc}
\cline{1-11}
\multirow{2}{*}{\textbf{Systems}} & \multicolumn{4}{c|}{\textbf{Ablation Setup}} & \multicolumn{6}{c}{\textbf{Emphasis Intensity Prediction Accuracy}} \\
\cline{2-11}
& CTE & MFTE & CAE & MFAE  & $\text{Match}_{1}$ ($\uparrow$) & $\text{Match}_{2}$ ($\uparrow$) & $\text{Match}_{mean}$ ($\uparrow$) &  $\text{F1}_{1}$ ($\uparrow$) & $\text{F1}_{2}$ ($\uparrow$) & $\text{F1}_{mean}$ ($\uparrow$) \\

\cline{1-11}
\textit{Abl.Exp.8} & {-}& {-}& {-}& {-}& 0.7014 & 0.8010 & 0.7512 & 0.5818 & 0.7046 &0.6432 \\
\textit{Abl.Exp.9} & $\checkmark$ &{-} & {-}& {-}& 0.7040 & 0.8025 & 0.7533 & 0.5858 & 0.7071 &0.6465  \\
\textit{Abl.Exp.10} &{-} & $\checkmark$ &{-} &{-} & 0.7095 & 0.8027 & 0.7561 & 0.5900 & 0.7059 &0.6480 \\
\textit{Abl.Exp.11} & $\checkmark$ & $\checkmark$ & {-}& {-}& 0.7069 & 0.8040 & 0.7555 & 0.5879 & 0.7077  &0.6478 \\
\textit{Abl.Exp.12} & {-}& {-}& $\checkmark$ & {-}& 0.7027 & 0.8034 & 0.7531 & 0.5831 & 0.7064  &0.6448  \\
\textit{Abl.Exp.13} &{-} &{-} &{-} & $\checkmark$ & 0.7107 & 0.8027 & 0.7567 & 0.5902 & 0.7062  &0.6482  \\
\textit{Abl.Exp.14} &{-} &{-} & $\checkmark$ & $\checkmark$ & 0.7099 & 0.8024 & 0.7562 & 0.5894 & 0.7058   &0.6476 \\
\textit{Abl.Exp.15} & $\checkmark$ &{-} & $\checkmark$ &{-} & 0.7078 & 0.8040 & 0.7559 & 0.5889 & 0.7079   &0.6484 \\
\textit{Abl.Exp.16} &{-} & $\checkmark$ &{-} & $\checkmark$ & 0.7082 & 0.8030 & 0.7556 & 0.5892 & 0.7073   &0.6483 \\
\cline{1-11}
\textbf{ER-CTTS} & $\checkmark$ & $\checkmark$ & $\checkmark$ & $\checkmark$ & \textbf{0.7116} & \textbf{0.8045} & \textbf{0.7581} & \textbf{0.5915} & \textbf{0.7079}  &\textbf{0.6497} \\
\cline{1-11}

\textit{Abl.Exp.17}: w/o Bidirectional Context Modeling & $\checkmark$ & $\checkmark$ & $\checkmark$ & $\checkmark$ & 0.7090 & 0.8026 & 0.7558 & 0.5896 & 0.7061 &0.6479   \\

\textit{Abl.Exp.18}: w/o Memory Enhancement &$\checkmark$ & $\checkmark$ & $\checkmark$ & $\checkmark$  &0.7031 & 0.8017 & 0.7524 & 0.5847 & 0.7058 &0.6453  \\

\textit{Abl.Exp.19}: w/o Hybrid-grained Fusion &$\checkmark$ & $\checkmark$ & $\checkmark$ & $\checkmark$  &0.7057 & 0.8021 & 0.7539 & 0.5862 & 0.7052 &0.6457  \\

\textit{Abl.Exp.20}: w/o Cross-modality Fusion &$\checkmark$ & $\checkmark$ & $\checkmark$ & $\checkmark$  &0.7073 & 0.8026 & 0.7550 & 0.5889 & 0.7057 &0.6473  \\
\cline{1-11}
\end{tabular}
}

\end{table*}

\section{Results and Discussions}
\subsection{Main Results}
The subjective and objective results of our ER-CTTS along with other baseline systems are reported in Table \ref{t-tab4}. For subjective evaluations, we invite 20 volunteers, including ten English major graduate students (5 male, 5 female) and ten speech processing major graduate students (5 male, 5 female), to participate in the listening experiments. They all have a strong command of the English language in terms of listening and speaking skills. Each volunteer provides individual ratings for a total of 140 synthesized samples, with each sample receiving a rating on a scale from 1 to 5.

From the first 8 rows of Table \ref{t-tab4}, it can be observed that ER-CTTS achieves the optimal performance in MAE-P (0.342), MAE-E (0.199), and MAE-D (0.148). However, for emphasis rendering, subjective performance is more reflective of human real feelings. In the subjective experiment, our proposed ER-CTTS achieves an N-DMOS score of 3.852 and an E-DMOS score of 3.893, surpassing all baseline models. ER-CTTS accurately extracts emphasis cues from dialogue history, infers the emphasized words in the current utterance based on the multi-modal and multi-scale knowledge within context, and thus synthesizes speech with notable emphasis intensity.

\subsection{Ablation Results}
We conduct ablation experiments from both emphasis rendering and prediction perspectives to validate the effectiveness and rationality of various modules in our ER-CTTS model.

\textbf{Emphasis Rendering}: The subjective and objective results are shown in rows 8-15 of Table \ref{t-tab4}. Note that Abl.Exp.1-4 are conducted to validate the effectiveness of the entire module, while Abl.Exp.5-7 verify the effectiveness of the corresponding key techniques within the module. It is observed that both subjective and objective results of the ablation experiments exhibit a significant decrease. Based on this, we draw the following three conclusions. 

\begin{itemize}
    \item \textbf{1) ER-CTTS effectively integrates information from multi-modal and multi-scale contexts:} Comparing ER-CTTS with Abl.Exp.1 and 2, the MOS scores of Exp show a significant decrease, indicating that both the coarse-grained and fine-grained encoders effectively extract multi-scale information from the conversation history. Furthermore, comparing ER-CTTS with Abl.Exp.3 and 4, all subjective and objective metrics also exhibit a noticeable decrease, demonstrating that our context fusion encoder successfully integrates multi-modal and multi-scale information for understanding the dialogue context and facilitating emphasis word inference.

    \item  \textbf{2) ER-CTTS fully leverages the inherent characteristics of dialogue to understand the emphasis cues in dialogue history:} Comparing ER-CTTS with Abl.Exp.5 and 6, our ER-CTTS also exhibits a clear advantage. This further demonstrates that ER-CTTS takes advantage of the bidirectional nature of information flow in dialogue and utilizes the emphasis information from the conversation history to enhance the overall information, thereby facilitating the inference of emphasis expression in the current sentence.

    \item  \textbf{3) The emphasis intensity information extracted from our designed emphasis annotation scheme strongly supports expressive emphasis rendering:} Comparing ER-CTTS with Abl.Exp.7, it is evident that emphasis features provide better emphasis expression capabilities compared to emphasis labels. This enables ER-CTTS to synthesize smoother speech with emphasized word expressions.
\end{itemize}

\textbf{Emphasis Prediction}: To provide a more objective and comprehensive analysis of the effectiveness of emphasis prediction in ER-CTTS, we conduct ablation experiments by systematically combining the four encoders, namely CTE, MFTE, CAE, and MFAE. Subsequently, we perform ablations on the bidirectional context modeling and memory enhancement mechanism. The objective results of the emphasis prediction are summarized in Table \ref{t-tab5}.

It is not surprising to find that Abl.Exp.8 achieves the lowest results in all metrics because it does not comprehensively model the conversation history. From the experimental results of Abl.Exp.10 vs. Abl.Exp.9 and Abl.Exp.13 vs. Abl.Exp.12, it can be observed that in both text and audio modalities, fine-grained encoding with memory enhancement performs better compared to coarse-grained encoders. From Abl.Exp.11 and Abl.Exp.14, we find that simply fusing multi-scale context of the single modality leads to a performance drop. By observing Abl.Exp.15 and Abl.Exp.16, we can conclude that solely fusing coarse-grained or fine-grained information alone cannot lead the model to achieve optimal performance. When all encoders are incorporated, the ER-CTTS model achieves optimal performance in all metrics. Additionally, by observing the last four rows of Table \ref{t-tab5}, all observations again demonstrate that the multi-scale and multi-modal context fusion mechanism fully understands the contextual information of dialogue history, infers the position and intensity of emphasis words in the current sentence, and thus achieves satisfactory emphasis prediction results.

\subsection{Context Length Analysis}
We also explore the effectiveness of emphasis modeling with different context lengths. Specifically, considering the average number of 9.3 dialogue turns in DailyTalk, we set the utterance length of dialogue history ranging from 4 to 16 to compare the objective performance of emphasis prediction. As shown in Table \ref{t-tab6}, the experiments demonstrate that the emphasis prediction achieves optimal performance when the context length is 10. This indicates that either insufficient or redundant context information can interfere with the model’s ability to understand emphasis cues in the context.

\begin{table}[h]
\centering
\caption{Objective results of various context lengths.}

\resizebox{0.5\textwidth}{!}{
\begin{tabular}{c|cccccc}
\hline
\textbf{Length} & $\text{Match}_{1}$ ($\uparrow$) & $\text{Match}_{2}$ ($\uparrow$) & $\text{Match}_{mean}$ ($\uparrow$) &  $\text{F1}_{1}$ ($\uparrow$) & $\text{F1}_{2}$ ($\uparrow$) & $\text{F1}_{mean}$ ($\uparrow$) \\
\hline
4 & 0.7082 & 0.8012 & 0.7547 & 0.5894 & 0.7049 & 0.6472 \\
6 & 0.7069 & 0.8020 & 0.7545 & 0.5877 & 0.7060 & 0.6469 \\
8 & 0.7073 & 0.8020 & 0.7547 & 0.5886 & 0.7057 & 0.6472 \\
\textbf{10} & \textbf{0.7116} & \textbf{0.8045} & \textbf{0.7581} & \textbf{0.5915} & \textbf{0.7079} & \textbf{0.6497} \\
12 & 0.7057 & 0.8033 & 0.7545 & 0.5868 & 0.7067 & 0.6468 \\
14 & 0.7111 & 0.8024 & 0.7568 & \textbf{0.5915} & 0.7060 & 0.6488 \\
16 & 0.7090 & 0.8034 & 0.7562 & 0.5913 & 0.7067 & 0.6490 \\
\hline
\end{tabular}
}

\label{t-tab6}
\end{table}

\begin{figure*}
  \centering
  \includegraphics[width=1\linewidth]{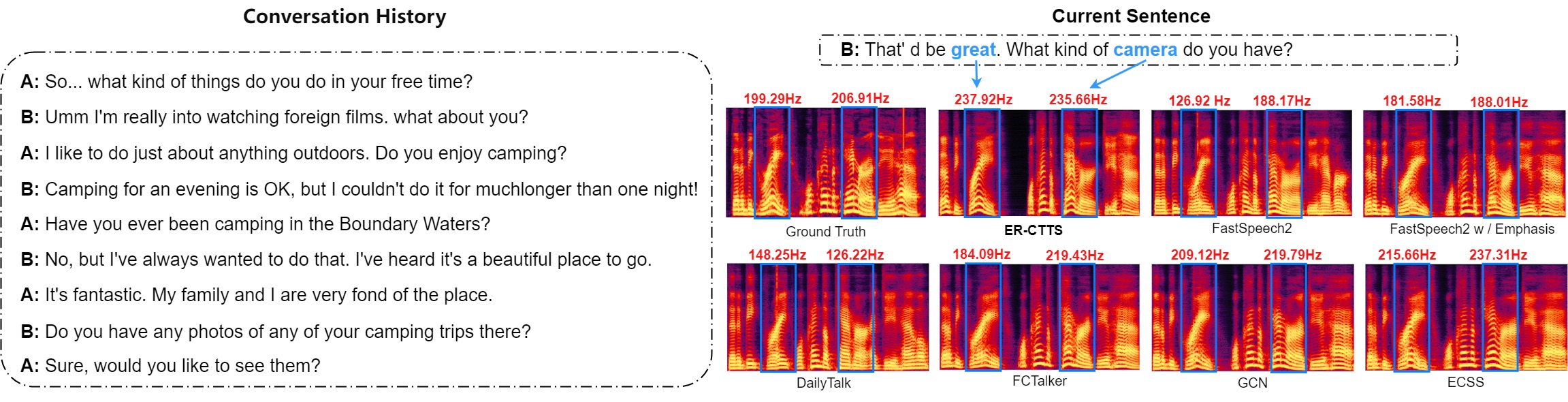}
  \caption{The mel-spectrogram and F0 plots of synthesized speech from different TTS systems. (The pitch value was separately computed. Blue boxes indicate annotated emphasis positions. The red font indicates the top F0 of the emphasized words that represent the emphasis intensity.)}
  \label{fig:5}
\end{figure*}

\subsection{Visualization Study}
To visually demonstrate the emphasis expressiveness of our ER-CTTS, we plot the Mel spectrograms of the synthesized speech samples for all systems and the Ground Truth. As shown in Fig. \ref{fig:5}, the blue boxes indicate the emphasis positions ``great'' and ``camera''. We can observe that the synthesized speech by ER-CTTS can restore more detailed spectral structures at these two positions. Additionally, it exhibits better rendering effects in terms of pitch, with higher pitch values at the two emphasized word positions compared to all other baselines. This demonstrates that our method can fully understand the context, enabling more accurate emphasis rendering in the current sentence. More speech samples are available at: \url{https://codestoretts.github.io/ER-CTTS/}.

\section{Conclusion}
To improve the emphasis rendering capability of CTTS systems, this paper proposes a novel ER-CTTS model. Our model can adequately model multi-modal and multi-scale contexts while mining implicit emphasis cues in the conversation history. It fully learns the underlying relationship between the conversation history and the emphasis expression in the current sentence, ultimately achieving accurate emphasis information inference and synthesizing expressive speech with emphasis rendering. Experimental results demonstrate the superiority of ER-CTTS over state-of-the-art CTTS systems. To the best of our knowledge, ER-CTTS is the first in-depth conversational speech synthesis study that models emphasis expressiveness. We hope that our work will serve as a basis for future intelligent HCI studies.

\section{Limitations}
One limitation of our work is that during the inference of emphasis information in the current sentence, we assume that the emphasis information in the conversation history is known. In the training stage, our data provides emphasis labels for the entire conversation. However, it is important to note that in the actual deployment of the model, we can initially predict emphasis words using single-sentence information of the first sentence that represents the conversation start. From the second sentence onwards, we can use the ER-CTTS method to predict the emphasis information in the current sentence.
Another limitation is that our model does not utilize visual modality information. Factors such as the speaker's facial expressions can also influence the emphasis expression in dialogue \cite{poria2019meld}. This aspect will be considered for further research.
\textcolor{black}{In addition, the focus of our work is on accurately predicting emphasized words in the current sentence using contextual information, rather than modeling the spontaneous phenomenon during conversational speech generation. This aspect also will be considered for further research to further improve the naturalness of CTTS.}
Lastly, it is important to note that our method focuses solely on the effectiveness of emphasis rendering. Regarding speech quality, with the advancements in TTS systems trained on large-scale datasets \cite{2024base}, the duration of our Emp-DailyTak dataset may still be considered limited. Therefore, in the future, we plan to collect larger-scale high-fidelity datasets or adopt an effective pretraining strategy with found data to achieve good performance in both speech quality and emphasis rendering.




\bibliographystyle{IEEEtran}
\bibliography{refs}

\begin{thebibliography}{10}
\providecommand{\url}[1]{#1}
\csname url@samestyle\endcsname
\providecommand{\newblock}{\relax}
\providecommand{\bibinfo}[2]{#2}
\providecommand{\BIBentrySTDinterwordspacing}{\spaceskip=0pt\relax}
\providecommand{\BIBentryALTinterwordstretchfactor}{4}
\providecommand{\BIBentryALTinterwordspacing}{\spaceskip=\fontdimen2\font plus
\BIBentryALTinterwordstretchfactor\fontdimen3\font minus \fontdimen4\font\relax}
\providecommand{\BIBforeignlanguage}[2]{{%
\expandafter\ifx\csname l@#1\endcsname\relax
\typeout{** WARNING: IEEEtran.bst: No hyphenation pattern has been}%
\typeout{** loaded for the language `#1'. Using the pattern for}%
\typeout{** the default language instead.}%
\else
\language=\csname l@#1\endcsname
\fi
#2}}
\providecommand{\BIBdecl}{\relax}
\BIBdecl

\bibitem{guo2021conversational}
H.~Guo, S.~Zhang, F.~K. Soong, L.~He, and L.~Xie, ``Conversational end-to-end tts for voice agents,'' in \emph{2021 IEEE Spoken Language Technology Workshop (SLT)}.\hskip 1em plus 0.5em minus 0.4em\relax IEEE, 2021, pp. 403--409.

\bibitem{tulshan2019survey}
A.~S. Tulshan and S.~N. Dhage, ``Survey on virtual assistant: Google assistant, siri, cortana, alexa,'' in \emph{Advances in Signal Processing and Intelligent Recognition Systems: 4th International Symposium SIRS 2018, Bangalore, India, September 19--22, 2018, Revised Selected Papers 4}.\hskip 1em plus 0.5em minus 0.4em\relax Springer, 2019, pp. 190--201.

\bibitem{seaborn2021voice}
K.~Seaborn, N.~P. Miyake, P.~Pennefather, and M.~Otake-Matsuura, ``Voice in human--agent interaction: A survey,'' \emph{ACM Computing Surveys (CSUR)}, vol.~54, no.~4, pp. 1--43, 2021.

\bibitem{mctear2022conversational}
M.~McTear, ``Conversational ai: Dialogue systems, conversational agents, and chatbots,'' \emph{Synthesis Lectures on Human Language Technologies}, vol.~13, no.~3, pp. 1--251, 2020.

\bibitem{liu2021controllable}
L.~Liu, J.~Hu, Z.~Wu, S.~Yang, S.~Yang, J.~Jia, and H.~Meng, ``Controllable emphatic speech synthesis based on forward attention for expressive speech synthesis,'' in \emph{2021 IEEE Spoken Language Technology Workshop (SLT)}.\hskip 1em plus 0.5em minus 0.4em\relax IEEE, 2021, pp. 410--414.

\bibitem{seshadri2021emphasis}
S.~Seshadri, T.~Raitio, D.~Castellani, and J.~Li, ``Emphasis control for parallel neural tts,'' \emph{arXiv preprint arXiv:2110.03012}, 2021.

\bibitem{raitio2022hierarchical}
T.~Raitio, J.~Li, and S.~Seshadri, ``Hierarchical prosody modeling and control in non-autoregressive parallel neural tts,'' in \emph{ICASSP 2022-2022 IEEE International Conference on Acoustics, Speech and Signal Processing (ICASSP)}.\hskip 1em plus 0.5em minus 0.4em\relax IEEE, 2022, pp. 7587--7591.

\bibitem{zhong2023ee}
Y.~Zhong, C.~Zhang, X.~Liu, C.~Sun, W.~Deng, H.~Hu, and Z.~Sun, ``Ee-tts: Emphatic expressive tts with linguistic information,'' \emph{arXiv preprint arXiv:2305.12107}, 2023.

\bibitem{chi2023multi}
W.~Chi, X.~Feng, L.~Xue, Y.~Chen, L.~Xie, and Z.~Li, ``Multi-granularity semantic and acoustic stress prediction for expressive tts,'' in \emph{2023 Asia Pacific Signal and Information Processing Association Annual Summit and Conference (APSIPA ASC)}.\hskip 1em plus 0.5em minus 0.4em\relax IEEE, 2023, pp. 2409--2415.

\bibitem{10379131}
R.~Liu, Y.~Hu, H.~Zuo, Z.~Luo, L.~Wang, and G.~Gao, ``Text-to-speech for low-resource agglutinative language with morphology-aware language model pre-training,'' \emph{IEEE/ACM Transactions on Audio, Speech, and Language Processing}, vol.~32, pp. 1075--1087, 2024.

\bibitem{10487819}
R.~Liu, B.~Sisman, G.~Gao, and H.~Li, ``Controllable accented text-to-speech synthesis with fine and coarse-grained intensity rendering,'' \emph{IEEE/ACM Transactions on Audio, Speech, and Language Processing}, vol.~32, pp. 2188--2201, 2024.

\bibitem{cole2014listening}
J.~Cole, T.~Mahrt, and J.~I. Hualde, ``Listening for sound, listening for meaning: Task effects on prosodic transcription.''

\bibitem{cole2017crowd}
J.~Cole, T.~Mahrt, and J.~Roy, ``Crowd-sourcing prosodic annotation,'' \emph{Computer Speech \& Language}, vol.~45, pp. 300--325, 2017.

\bibitem{morrison2024crowdsourced}
M.~Morrison, P.~Pawar, N.~Pruyne, J.~Cole, and B.~Pardo, ``Crowdsourced and automatic speech prominence estimation,'' in \emph{ICASSP 2024-2024 IEEE International Conference on Acoustics, Speech and Signal Processing (ICASSP)}.\hskip 1em plus 0.5em minus 0.4em\relax IEEE, 2024, pp. 12\,281--12\,285.

\bibitem{he2022automatic}
W.~He, Y.~Lin, J.~Ye, H.~Zhou, K.~Ren, T.~He, P.~Tan, and H.~Lu, ``Automatic stress annotation and prediction for expressive mandarin tts,'' in \emph{National Conference on Man-Machine Speech Communication}.\hskip 1em plus 0.5em minus 0.4em\relax Springer, 2022, pp. 306--317.

\bibitem{xue2023m}
J.~Xue, Y.~Deng, F.~Wang, Y.~Li, Y.~Gao, J.~Tao, J.~Sun, and J.~Liang, ``M 2-ctts: End-to-end multi-scale multi-modal conversational text-to-speech synthesis,'' in \emph{ICASSP 2023-2023 IEEE International Conference on Acoustics, Speech and Signal Processing (ICASSP)}.\hskip 1em plus 0.5em minus 0.4em\relax IEEE, 2023, pp. 1--5.

\bibitem{lei2023msstyletts}
S.~Lei, Y.~Zhou, L.~Chen, Z.~Wu, X.~Wu, S.~Kang, and H.~Meng, ``Msstyletts: Multi-scale style modeling with hierarchical context information for expressive speech synthesis,'' \emph{IEEE/ACM Transactions on Audio, Speech, and Language Processing}, 2023.

\bibitem{shang2023hiertts}
Z.~Shang, P.~Shi, P.~Zhang, L.~Wang, and G.~Zhao, ``Hiertts: Expressive end-to-end text-to-waveform using a multi-scale hierarchical variational auto-encoder,'' \emph{Applied Sciences}, vol.~13, no.~2, p. 868, 2023.

\bibitem{hu2022fctalker}
Y.~Hu, R.~Liu, G.~Gao, and H.~Li, ``Fctalker: Fine and coarse grained context modeling for expressive conversational speech synthesis,'' \emph{arXiv preprint arXiv:2210.15360}, 2022.

\bibitem{lee2023dailytalk}
K.~Lee, K.~Park, and D.~Kim, ``Dailytalk: Spoken dialogue dataset for conversational text-to-speech,'' in \emph{ICASSP 2023-2023 IEEE International Conference on Acoustics, Speech and Signal Processing (ICASSP)}.\hskip 1em plus 0.5em minus 0.4em\relax IEEE, 2023, pp. 1--5.

\bibitem{li2022enhancing}
J.~Li, Y.~Meng, C.~Li, Z.~Wu, H.~Meng, C.~Weng, and D.~Su, ``Enhancing speaking styles in conversational text-to-speech synthesis with graph-based multi-modal context modeling,'' in \emph{ICASSP 2022-2022 IEEE International Conference on Acoustics, Speech and Signal Processing (ICASSP)}.\hskip 1em plus 0.5em minus 0.4em\relax IEEE, 2022, pp. 7917--7921.

\bibitem{li2022inferring}
J.~Li, Y.~Meng, X.~Wu, Z.~Wu, J.~Jia, H.~Meng, Q.~Tian, Y.~Wang, and Y.~Wang, ``Inferring speaking styles from multi-modal conversational context by multi-scale relational graph convolutional networks,'' in \emph{Proceedings of the 30th ACM International Conference on Multimedia}, 2022, pp. 5811--5820.

\bibitem{liu2023emotion}
R.~Liu, Y.~Hu, Y.~Ren, X.~Yin, and H.~Li, ``Emotion rendering for conversational speech synthesis with heterogeneous graph-based context modeling,'' in \emph{Proceedings of the AAAI Conference on Artificial Intelligence}, 2024, pp. 1--9.

\bibitem{busso2008iemocap}
C.~Busso, M.~Bulut, C.-C. Lee, A.~Kazemzadeh, E.~Mower, S.~Kim, J.~N. Chang, S.~Lee, and S.~S. Narayanan, ``Iemocap: Interactive emotional dyadic motion capture database,'' \emph{Language resources and evaluation}, vol.~42, pp. 335--359, 2008.

\bibitem{poria2019meld}
S.~Poria, D.~Hazarika, N.~Majumder, G.~Naik, E.~Cambria, and R.~Mihalcea, ``Meld: A multimodal multi-party dataset for emotion recognition in conversations,'' in \emph{Proceedings of the 57th Annual Meeting of the Association for Computational Linguistics}.\hskip 1em plus 0.5em minus 0.4em\relax Association for Computational Linguistics, 2019.

\bibitem{zhao2022m3ed}
J.~Zhao, T.~Zhang, J.~Hu, Y.~Liu, Q.~Jin, X.~Wang, and H.~Li, ``M3ed: Multi-modal multi-scene multi-label emotional dialogue database,'' in \emph{Proceedings of the 60th Annual Meeting of the Association for Computational Linguistics (Volume 1: Long Papers)}, 2022, pp. 5699--5710.

\bibitem{yang2022open}
Z.~Yang, Y.~Chen, L.~Luo, R.~Yang, L.~Ye, G.~Cheng, J.~Xu, Y.~Jin, Q.~Zhang, P.~Zhang \emph{et~al.}, ``Open source magicdata-ramc: A rich annotated mandarin conversational (ramc) speech dataset,'' \emph{arXiv preprint arXiv:2203.16844}, 2022.

\bibitem{yang2019xlnet}
Z.~Yang, Z.~Dai, Y.~Yang, J.~Carbonell, R.~R. Salakhutdinov, and Q.~V. Le, ``Xlnet: Generalized autoregressive pretraining for language understanding,'' \emph{Advances in neural information processing systems}, vol.~32, 2019.

\bibitem{ghosal2019dialoguegcn}
D.~Ghosal, N.~Majumder, S.~Poria, N.~Chhaya, and A.~Gelbukh, ``Dialoguegcn: A graph convolutional neural network for emotion recognition in conversation,'' in \emph{Proceedings of the 2019 Conference on Empirical Methods in Natural Language Processing and the 9th International Joint Conference on Natural Language Processing (EMNLP-IJCNLP)}, 2019, pp. 154--164.

\bibitem{reimers2019sentence}
N.~Reimers and I.~Gurevych, ``Sentence-bert: Sentence embeddings using siamese bert-networks,'' in \emph{Proceedings of the 2019 Conference on Empirical Methods in Natural Language Processing and the 9th International Joint Conference on Natural Language Processing (EMNLP-IJCNLP)}, 2019, pp. 3982--3992.

\bibitem{liu2023emotionally}
Y.~Liu, H.~Zhang, S.~Liu, X.~Yin, Z.~Ma, and Q.~Jin, ``Emotionally situated text-to-speech synthesis in user-agent conversation,'' in \emph{Proceedings of the 31st ACM International Conference on Multimedia}, 2023, pp. 5966--5974.

\bibitem{wu-etal-2020-tod}
\BIBentryALTinterwordspacing
C.-S. Wu, S.~C. Hoi, R.~Socher, and C.~Xiong, ``{TOD}-{BERT}: Pre-trained natural language understanding for task-oriented dialogue,'' in \emph{Proceedings of the 2020 Conference on Empirical Methods in Natural Language Processing (EMNLP)}, B.~Webber, T.~Cohn, Y.~He, and Y.~Liu, Eds.\hskip 1em plus 0.5em minus 0.4em\relax Online: Association for Computational Linguistics, Nov. 2020, pp. 917--929. [Online]. Available: \url{https://aclanthology.org/2020.emnlp-main.66}
\BIBentrySTDinterwordspacing

\bibitem{baevski2020wav2vec}
A.~Baevski, Y.~Zhou, A.~Mohamed, and M.~Auli, ``wav2vec 2.0: A framework for self-supervised learning of speech representations,'' \emph{Advances in neural information processing systems}, vol.~33, pp. 12\,449--12\,460, 2020.

\bibitem{ren2020fastspeech}
Y.~Ren, C.~Hu, X.~Tan, T.~Qin, S.~Zhao, Z.~Zhao, and T.-Y. Liu, ``Fastspeech 2: Fast and high-quality end-to-end text to speech,'' \emph{arXiv preprint arXiv:2006.04558}, 2020.

\bibitem{mass2018word}
Y.~Mass, S.~Shechtman, M.~Mordechay, R.~Hoory, O.~{Sar Shalom}, G.~Lev, and D.~Konopnicki, ``{Word Emphasis Prediction for Expressive Text to Speech},'' in \emph{Proc. Interspeech 2018}, 2018, pp. 2868--2872.

\bibitem{sang2003introduction}
E.~T.~K. Sang and F.~De~Meulder, ``Introduction to the conll-2003 shared task: Language-independent named entity recognition,'' in \emph{Proceedings of the Seventh Conference on Natural Language Learning at HLT-NAACL 2003}, 2003, pp. 142--147.

\bibitem{shirani2019learning}
A.~Shirani, F.~Dernoncourt, P.~Asente, N.~Lipka, S.~Kim, J.~Echevarria, and T.~Solorio, ``Learning emphasis selection for written text in visual media from crowd-sourced label distributions,'' in \emph{Proceedings of the 57th Annual Meeting of the Association for Computational Linguistics}, 2019, pp. 1167--1172.

\bibitem{fleiss2013statistical}
J.~L. Fleiss, B.~Levin, and M.~C. Paik, \emph{Statistical methods for rates and proportions}.\hskip 1em plus 0.5em minus 0.4em\relax john wiley \& sons, 2013.

\bibitem{chen2019emotionlines}
S.~Y. Chen, C.~C. Hsu, C.~C. Kuo, T.~H.~K. Huang, and L.~W. Ku, ``Emotionlines: An emotion corpus of multi-party conversations,'' in \emph{11th International Conference on Language Resources and Evaluation, LREC 2018}.\hskip 1em plus 0.5em minus 0.4em\relax European Language Resources Association (ELRA), 2019, pp. 1597--1601.

\bibitem{streijl2016mean}
R.~C. Streijl, S.~Winkler, and D.~S. Hands, ``Mean opinion score (mos) revisited: methods and applications, limitations and alternatives,'' \emph{Multimedia Systems}, vol.~22, no.~2, pp. 213--227, 2016.

\bibitem{mcauliffe17_interspeech}
M.~McAuliffe, M.~Socolof, S.~Mihuc, M.~Wagner, and M.~Sonderegger, ``{Montreal Forced Aligner: Trainable Text-Speech Alignment Using Kaldi},'' in \emph{Proc. Interspeech 2017}, 2017, pp. 498--502.

\bibitem{2024base}
M.~Łajszczak, G.~Cámbara, Y.~Li, F.~Beyhan, A.~van Korlaar, F.~Yang, A.~Joly, Álvaro Martín-Cortinas, A.~Abbas, A.~Michalski, A.~Moinet, S.~Karlapati, E.~Muszyńska, H.~Guo, B.~Putrycz, S.~L. Gambino, K.~Yoo, E.~Sokolova, and T.~Drugman, ``Base tts: Lessons from building a billion-parameter text-to-speech model on 100k hours of data,'' \emph{arXiv preprint arXiv:2402.08093}, 2024.

\end{thebibliography}

\vfill

\end{document}